\documentclass{article}

\usepackage{PRIMEarxiv}

\usepackage[utf8]{inputenc} 
\usepackage[T1]{fontenc}    
\usepackage{hyperref}       
\usepackage{url}            
\usepackage{booktabs}       
\usepackage{amsfonts}       
\usepackage{nicefrac}       
\usepackage{microtype}      
\usepackage{lipsum}
\usepackage{fancyhdr}       
\usepackage{graphicx}       
\graphicspath{{media/}}     
\usepackage{subcaption}
\usepackage{wrapfig}

\title{Nadine: An LLM-driven Intelligent Social Robot with Affective Capabilities and Human-like Memory
}

\author{
  Hangyeol Kang$^*$, {} {} {} Maher Ben Moussa$^*$, {} {} {} Nadia Magnenat-Thalmann \\
  Centre Universitaire d’Informatique \\
  University of Geneva \\
  Geneva, Switzerland\\
  \texttt{\{hangyeol.kang, maher.benmoussa, nadia.thalmann\} @ unige.ch} \\
}

\begin{document}
\maketitle

\def\thefootnote{*}\footnotetext{Equal contribution.}\def\thefootnote{\arabic{footnote}}

\begin{abstract}
In this work, we describe our approach to developing an intelligent and robust social robotic system for the Nadine social robot platform. We achieve this by integrating Large Language Models (LLMs) and skilfully leveraging the powerful reasoning and instruction-following capabilities of these types of models to achieve advanced human-like affective and cognitive capabilities. This approach is novel compared to the current state-of-the-art LLM-based agents which do not implement human-like long-term memory or sophisticated emotional appraisal. The naturalness of social robots, consisting of multiple modules, highly depends on the performance and capabilities of each component of the system and the seamless integration of the components. We built a social robot system that enables generating appropriate behaviours through multimodal input processing, bringing episodic memories accordingly to the recognised user, and simulating the emotional states of the robot induced by the interaction with the human partner. In particular, we introduce an LLM-agent frame for social robots, SoR-ReAct, serving as a core component for the interaction module in our system. This design has brought forth the advancement of social robots and aims to increase the quality of human-robot interaction.

\end{abstract}


\section{Introduction}
\label{sec:1}

In recent decades, robotics research has expanded significantly, encompassing a wide array of applications spanning from industrial robotics \cite{buerkle2023towards,soori2023optimization} to social robotics\cite{thalmann2017nadine,alam2022social,naneva2020systematic}. In this work, we focus on social robots, specifically designed for human interaction, which have gained increasing attention due to their potential roles in healthcare, elderly care, and various service industries \cite{johal2020research,thalmann2021nadine,wu2023bloomberggpt} such as education, museum and finance. This surge in interest has led to extensive research endeavours aimed at enhancing the quality of human-robot interaction (HRI) by refining robot systems to mimic human-like behaviours \cite{ramanathan2019nadine,won2020adaptive,cruz2023explainable}, and exploring the impact of social attributes on HRI dynamics \cite{lazzeri2018,kasap2009making,mishra2022nadine} and the influence of the physical appearance of social robots \cite{kluber2022appearance,liu2022friendly,pinto2023different,ringwald2023should}.

\begin{figure}
 \centering
 \includegraphics[width=0.5\linewidth]{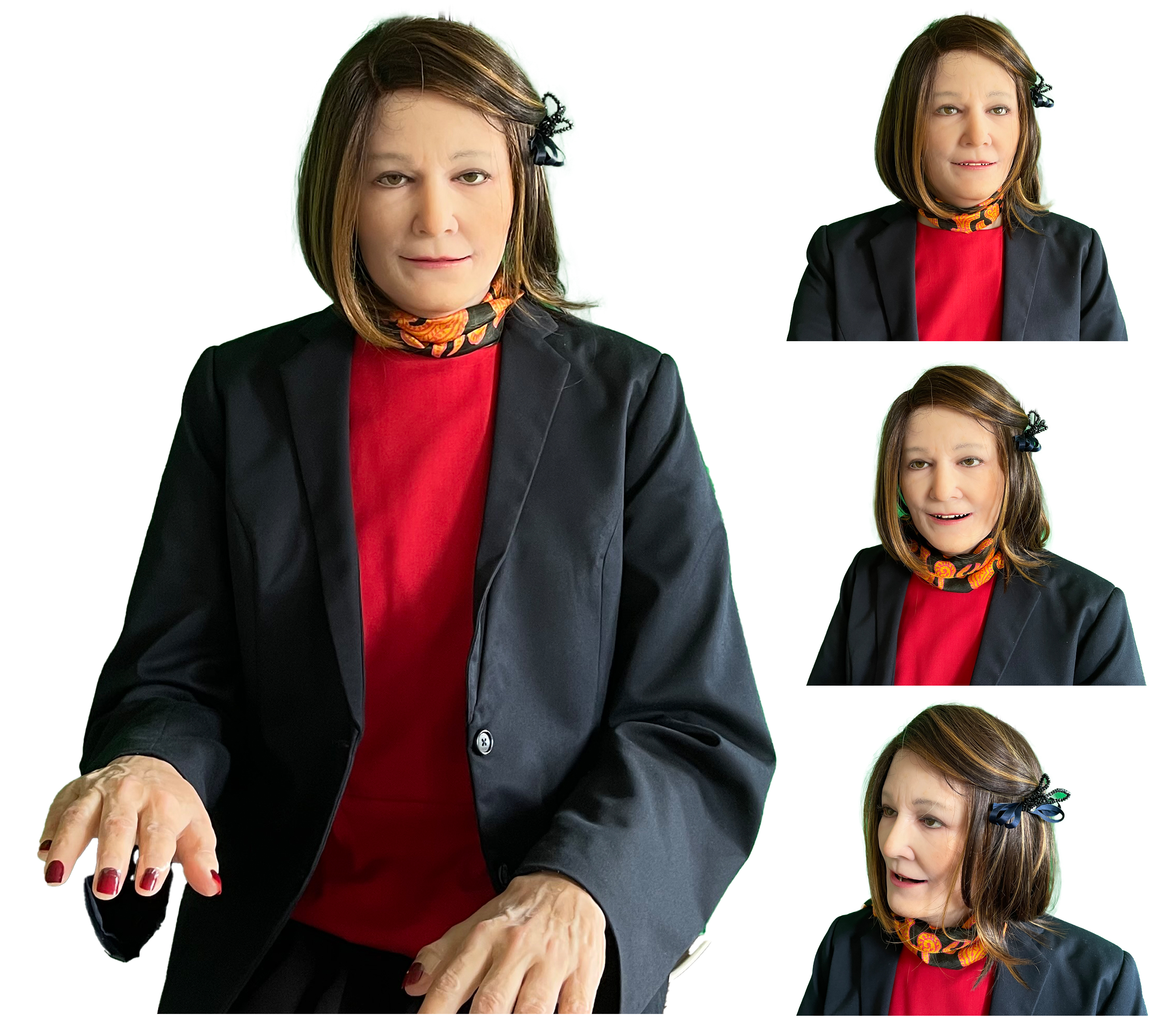}
 \caption{Nadine social robot new appearance.}
\label{fig:Trans_architecture}
\end{figure}

Large Language Models (LLMs) represent a pivotal advancement in artificial intelligence, trained on vast corpora of textual data, possessing the ability to comprehend intricate linguistic nuances, infer contextual meanings, and generate coherent responses. Their versatility has rendered them indispensable across a spectrum of industries, from healthcare to finance to entertainment. Notably, the integrations of LLMs and robotics have emerged as a transformative trend, offering new avenues for enhancing robot capabilities \ref{zeng2022socratic}.

This paper presents a novel robotics system deployed in the social robot Nadine, which comprises three key modules: perception, interaction, and robot control modules. The perception module takes the role of understanding multiple modalities encompassing the user’s query and the environmental visual cues. The data is passed to the interaction module and used to generate appropriate robot behaviours through multiple components such as the dialogue manager, retrieval augmented generation (RAG)-based memory and knowledge system, and affective system. Our approach aims to leverage the versatility of an LLM-powered agent to effectively process the multi-modal inputs for generating appropriate behaviours of the robot. In practice, the robot can recognise the user (once the user has already interacted with it in the past time), and recall the episodic memory. At the same time, the robot generates its emotional internal states and reflects the emotion in generating its behaviour. The affective system helps the robot establish better connections with people.

Our contributions can be summarised as follows: (1) We introduce SoR-ReAct, a novel LLM agent for the social robotic system employed in the Nadine social robot. (2) We demonstrate the robotic system for effective episodic memory of the recognised user. (3) We present a sophisticated affective system that generates the robot's emotions evoked from interactions with the user and detects the user's emotions, which are reflected through the robot's behaviours. Empirical observations revealed that these elements can help people experience a stronger connection with the robot.

The rest of the paper is organised as follows: Section \ref{sec:2} provides an overview of related work in the LLM for the robotics field. Section \ref{sec:3} introduces our LLM-powered agent-based robotic system deployed in the Nadine social robot. Section \ref{sec:4} discusses the noteworthy findings of our system via empirical observations. Finally, in section \ref{sec:5}, we present our conclusions and future work.

\section{Related Work}
\label{sec:2}

In the context of robotics, a multitude of research has been conducted on capitalising on the power of LLMs \cite{huang2022inner}. Certain research focuses on controlling robotic systems with natural language by leveraging the power of LLMs. LLM-BRAIn \cite{lykov2023llm} generates possible robot behaviour trees from text description using a fine-tuned Alpaca 7B model. LLM-MARS \cite{lykov2023llmmars} applies a multimodal LLM to control a tandem of mobile robots under conditions that closely emulate real-world scenarios. BTGenBot \cite{izzo2024btgenbot} showcases the efficiency of compact LLMs in generating executable behaviour trees tailored to robot behaviours. Another approach using the power of LLMs revolves around robot task planning. SayCan \cite{ahn2022can} leverages the substantial knowledge of LLMs by grounding them for robotic planning. ProgPrompt \cite{singh2023progprompt} presents a programmatic LLM prompt structure for generating plans across diverse environments and tasks by providing program-like specifications and executable examples. Wang et al. \cite{wang2024safe} and Kannan et al. \cite{kannan2023smart} leverage the power of LLMs to generate task plans for a group of robots. Researchers have also explored the potential of LLMs for human-robot interactions. Lim et al. \cite{lim2023sign} employ an LLM to enable their robot to generate natural Co-Speech Gesture responses, particularly embedded with a sign language recognition system. Spitale et al. \cite{spitale2023vita} leverage a multi-modal LLM-based system for robotic coaches. Kim et al. \cite{kim2024understanding} demonstrate that LLM-equipped robots enhance user expectations for non-verbal cues and excel in connection building and deliberation. Feng et al. \cite{feng2024large} introduce a reinforcement learning-based method for LLM-based human-agent collaboration for complex task-solving. Tanneberg et al. \cite{tanneberg2024help} leverage the common-sense reasoning capabilities of LLMs to make their robot aware of when and how to support humans. In addition to the aforementioned approaches, much research has been conducted on robotics utilising the potential of LLMs \cite{wu2023tidybot,karli2024alchemist}. Most of the mentioned implementations focus on leveraging reasoning and action planning capabilities to plan tasks and solve problems. In our SoR-ReAct implementation, we aim to harness the powerful capabilities of action planning and reasoning. Furthermore, to achieve more realistic human-like functionalities, we are also integrating personalized long-term memory and introducing emotional capabilities to LLMs.

\section{Nadine Framework}
\label{sec:3}

The interaction with a social humanoid robot requires equipping the robot with the capabilities of understanding the user’s speech and the environment, processing the multi-modalities, and generating natural responses, including gesture, facial expression, lip synchronisation, and speech. Each part plays a crucial role in developing a natural social humanoid robot. In this section, we explain the system architecture of our social robot, Nadine.

\begin{figure*}[ht]
 \includegraphics[width=\textwidth]{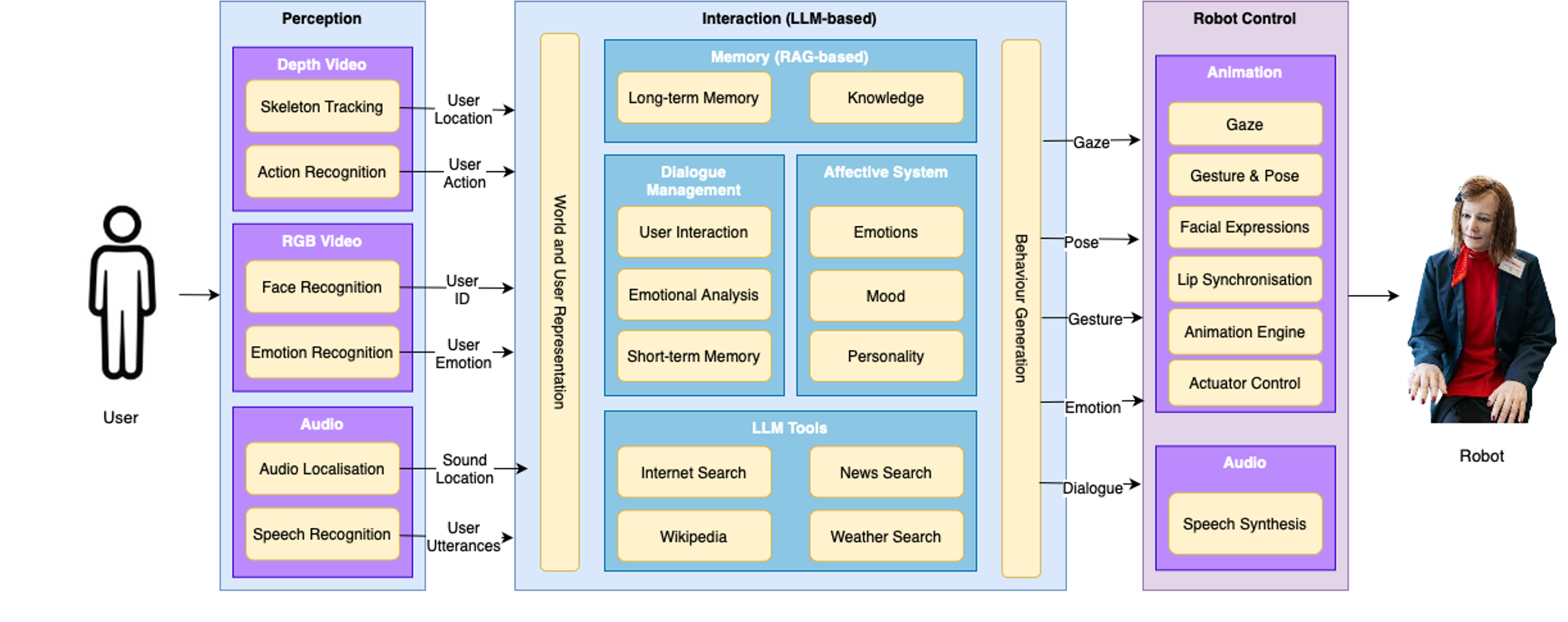}
 \caption{Nadine system architecture. The system comprises three key modules: the perception module, the interaction module, and the control module. This study specifically focuses on the interaction module, which equips a social robot with an LLM-powered agent to achieve human-like affective and cognitive capabilities.}
\label{fig:system_architecture}
\end{figure*}

\subsection{Overview of the system}
\label{sec:3.1}

The overall architecture of Nadine’s system is illustrated in Figure \ref{fig:system_architecture}. The system consists of three modules: perception module, interaction module, and robot control module. The perception module takes the role of understanding the user’s speech and the environment with sensors including an RGB camera, a depth camera and a microphone. Specifically, it tracks the user’s skeleton and recognises the user’s actions, face, and emotions. The perceived visual and auditory information is passed to the interaction module, which comprehensively processes the detected user representations. We adopt a Large Language Model (LLM) powered agent, called SoR-ReAct, to enable the robot to effectively process the information. The SoR-ReAct agent is equipped with a RAG-based memory system, a tool-use system and an affective system. Through this process, natural behaviours including gaze, pose, gesture, emotion, and dialogue are generated and transferred to the robot control module. The robot control module takes the generated robot’s behaviours and actuates the robot. The details of each module are described in the following sections.

\subsection{Perception Module}
\label{sec:3.2}
The perception module is crucial for smooth interaction with the robot as a starting point of the robot system. The information of the user and their environment is captured by sensors: a depth camera, an RGB camera, and a microphone. For the cameras, we utilise the Microsoft Kinect V2 3D camera.\footnote[1]{\url{https://www.microsoft.com/en-us/download/details.aspx?id=44561}} Leveraging the depth data captured by the 3D camera, the module tracks the user’s skeleton and recognises their actions. The interaction module harnesses the user’s skeleton in localising the user. In the meantime, the module localises the user’s face and recognises the user by the detected face. In the case that the user has already interacted with the robot in the past, the robot can recognise the user and the user’s ID is leveraged to recall the episodic memory of conversations that the user had with the robot before. Otherwise, the module captures a frame containing the user and assigns a new user ID to the new user. The newly captured frame is used to recognise the user in later interactions. This process is carried out automatically by comparing the detected faces and the stored faces in the database. We leverage the DeepFace\footnote[2]{\url{https://github.com/serengil/deepface}} framework for face recognition, a lightweight hybrid face recognition framework. Furthermore, the RGB camera enables the perception module to recognise the user’s emotions. The recognised emotions are transferred to the dialogue management system in the interaction module and Utilized to generate the appropriate responses. The user’s query is captured through a microphone and the perception module localises the audio and recognises the user's utterances. The localised audio is used for the interaction module to localise the user along with the detected user’s skeleton. In addition, the user's speech is transferred to the interaction module in the form of text transformed by a speech-to-text model from Google Cloud.\footnote[3]{\url{https://cloud.google.com/speech-to-text}}

\subsection{Interaction Module}
\label{sec:3.3}

\subsubsection{Dialogue management}
\label{sec:3.3.1}

\begin{figure*}[ht]
    \centering
    \includegraphics[width=1\textwidth]{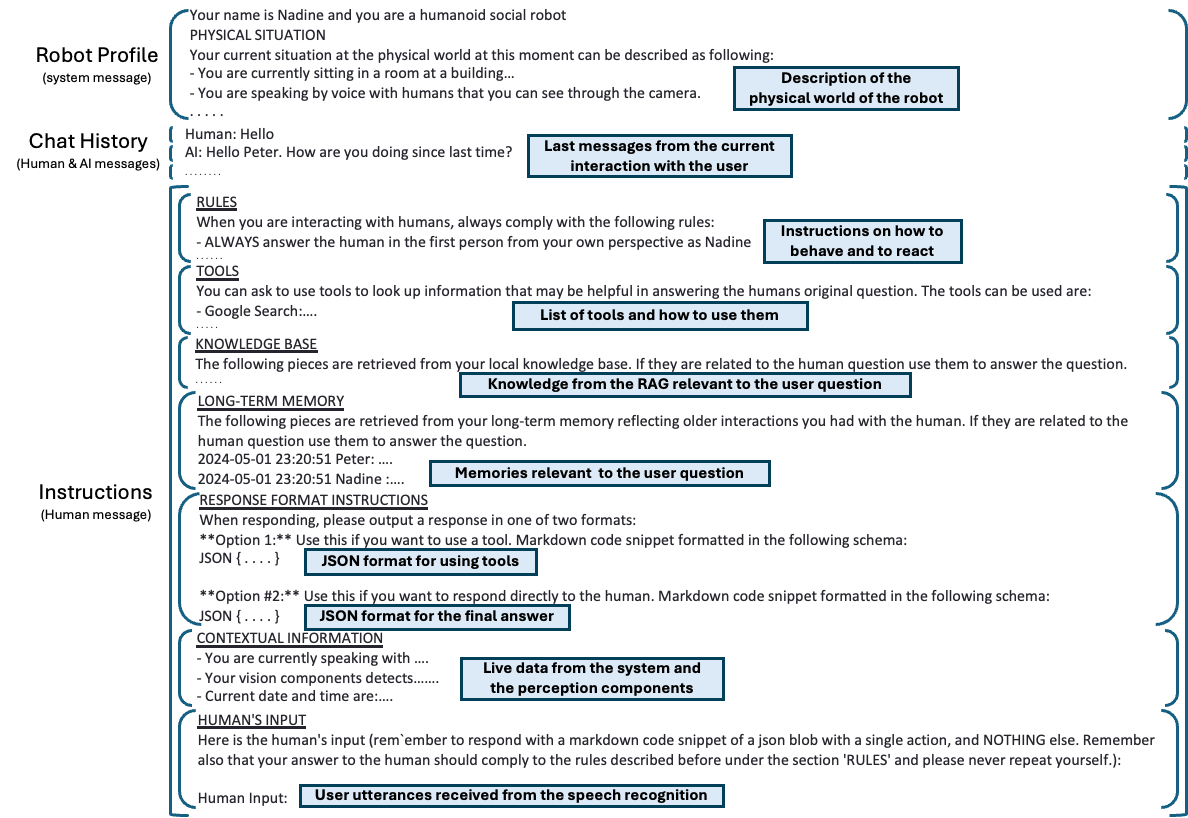}
    \caption{SoR-ReAct prompt structure}
    \label{fig:promptstructure}
\end{figure*}

Our dialogue management module serves as the core component for managing the interaction with the user and with the robot’s cognitive and affective components. Its architecture, which we named SoR-ReAct, enhances the ReAct framework\cite{yao2022react} by integrating human-like cognitive and affective capabilities to enable coherent and context-aware social interactions with users. ReAct, which stands for Reasoning and Acting, is a well-established agent framework for LLMs. It was one of the first frameworks to incorporate prompting for generating reasoning traces (chain-of-thought, or CoT) and task-specific action generation in an interleaved manner. This approach aims to reduce LLM hallucinations and incorporate data from external sources and tools into user responses.\\

\begin{figure}[h]
    \centering
    \includegraphics[width=0.7\textwidth]{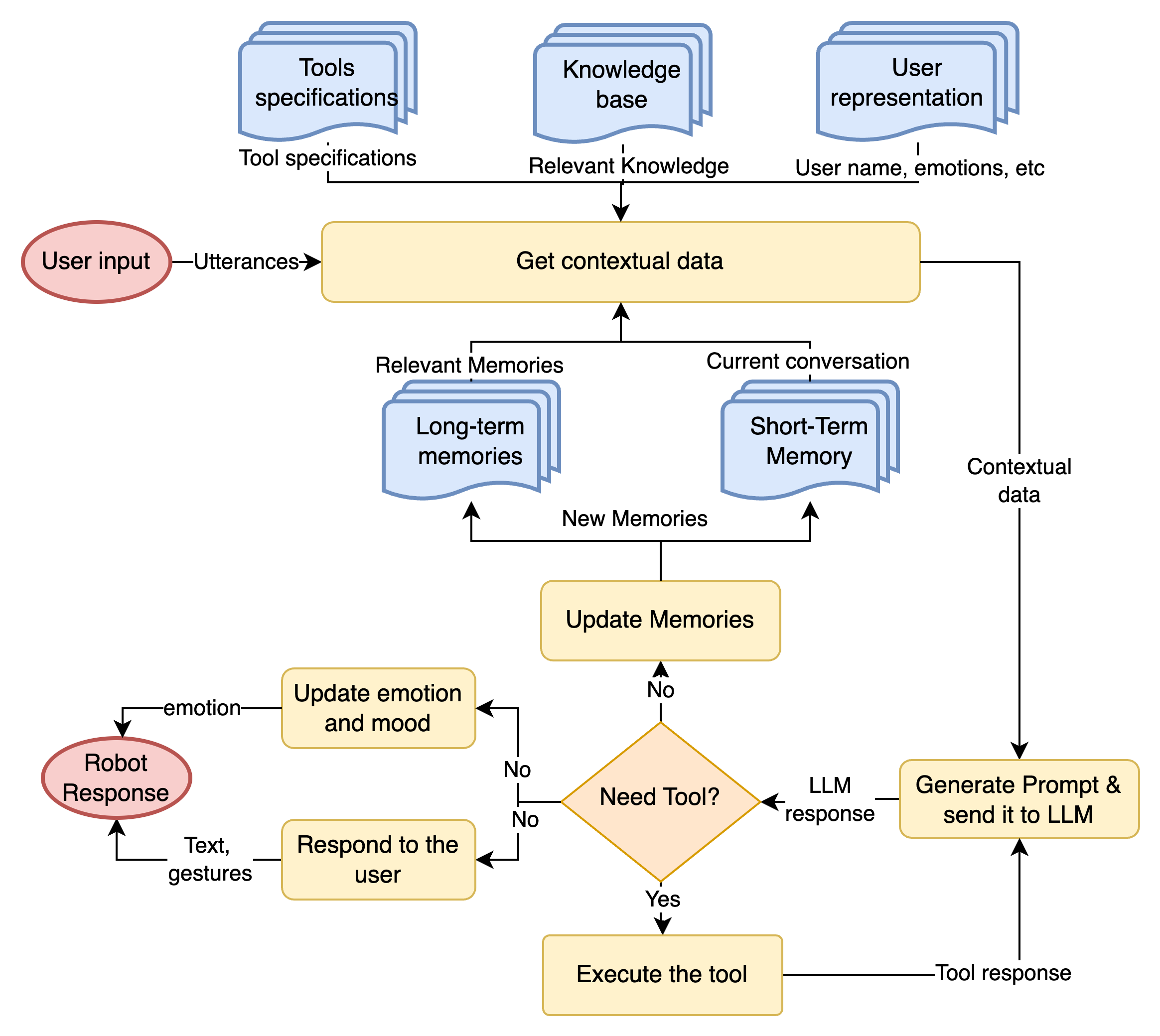}
    \caption{SoR-ReAct workflow}
    \label{fig:dm_workflow}
\end{figure}
Our approach of SoR-ReAct follows the same principles of an interleaved interaction between reasoning and action generation and implements an advanced prompting mechanism that integrates several tools that give the robot access to external information. However, contrary to the traditional ReAct agent that is mainly intended for simple question answering and information enquiry, SoR-ReAct aims to endow robots with the ability to function as a genuine social companion. This includes: (1) holding previous interactions with multiple users and using the long-term memories in future interactions when relevant, (2) possessing specialized knowledge pertinent to its role (such as hotel receptionist, psychologist, or bank employee) and responding to users based on this knowledge and (3) exhibiting a distinct personality and responding based on that personality, influenced by emotions arising from interactions with users.\\
Based on those characteristics, a sophisticated prompt (Figure \ref{fig:promptstructure}) is developed together with an advanced mechanism (Figure \ref{fig:dm_workflow}) that next to the ReAct tools functionality also integrates knowledge, memory, perception, and emotions. Due to the utilisation of GPT-4\cite{achiam2023gpt}, which is considered a “chat” model by OpenAI, the prompt is not one block but is divided into several blocks with each block representing one message. As illustrated in Figure \ref{fig:promptstructure}, the generated prompt consists of a list of messages.  
\\

\underline{\textbf{The SoR-ReAct workflow}}  
\\
As illustrated in Figure \ref{fig:dm_workflow}, the workflow of SoR-ReAct consists of the following steps:
\\
\textbf{\textit{Step 1:}} Each time user speech is detected, the following information will be retrieved:
\begin{itemize}
\item Long-term memories that are relevant to the user question and the current conversation
\item Knowledge that is relevant to the user question and the current conversation
\item Messages from the current conversation with the user (short-term memory)
\item The current user-specific data (user identity, detected user emotions, etc) from the perception components that can influence the response to the user utterances
\item Specifications of the active tools
\end{itemize}

\textbf{\textit{Step 2:}} Based on the retrieved information, a prompt is generated based on the specifications illustrated in Figure \ref{fig:promptstructure}, where short-term memory generates chat history, retrieved knowledge, long-term memories, user-related data and tools specifications are used to generate the corresponding blocks in the “Instructions” message. Following the ReAct approach, SoR-ReAct basically provides the generated prompts to the LLM to handle the response. The LLM can request one of the tools if it does not know the answer to the user's question, otherwise, it will return a response to the user.
\\
\textbf{\textit{Step 3:}} In a manner akin to the ReAct agent, when the LLM encounters a question it cannot answer, it will identify the appropriate tool required to address the query and provide the necessary input values for that tool. Consequently, SoR-ReAct would execute that tool, which can be for example a search engine or a weather app, and provide the response of that tool to the LLM. This process of which an example prompt is illustrated below can be continuously repeated until the user's answer is answered. 
\begin{figure}[h]
    \centering
    \begin{subfigure}[b]{0.45\textwidth}
        \centering
        \includegraphics[width=1\textwidth]{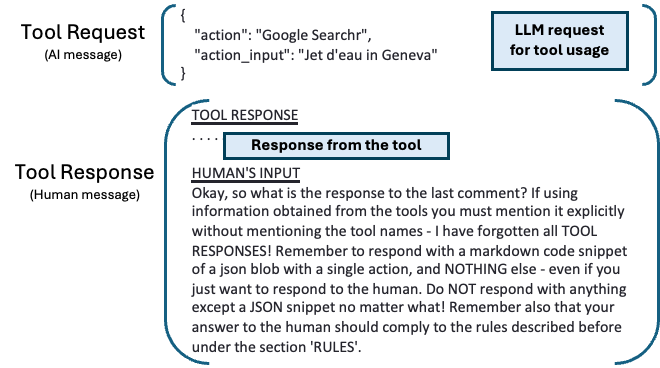}
        \caption{ }
        \label{fig:prompt_tools}
    \end{subfigure}
    \begin{subfigure}[b]{0.45\textwidth}
        \centering
        \includegraphics[width=1\textwidth]{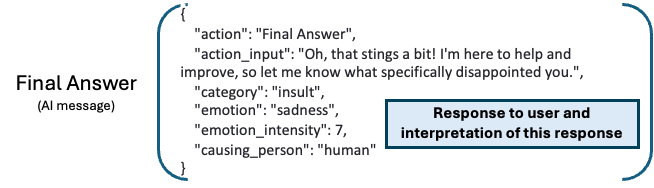}
        \caption{ }
        \label{fig:prompt_final}
    \end{subfigure}
    \caption{(a): LLM tools request and response. (b): LLM final response.}
\end{figure}

\noindent \textbf{\textit{Step 4:}} If the LLM is able to answer the user’s question based on its own knowledge or based on the knowledge already provided in the prompt, it will reply as in the following example in Figure \ref{fig:prompt_final}


The generated response will be used to update the memories, update the emotions and mood and generate user speech, gestures and facial expressions via the robot control module.
The response follows a well-defined template where the LLM analyses the user input, the current memory and the robot response and determines the fitting robot emotion with this context. A choice between the six basic emotions or a neutral emotion. Furthermore, the LLM is also instructed to determine the intensity of this emotion and the person that may have caused the emotion. The emotion category, the intensity, and the causing person are subsequently passed to the affective system to interact with mood and personality and to determine the appropriate emotional response of the robot. 
Additionally, the category (greeting, insult, compliment, etc) of the interaction is also determined by the LLM and is subsequently passed to the robot control module to generate the corresponding gestures.

\subsubsection{Affective system}
\label{sec:3.3.2}

\begin{wrapfigure}{r}{0.5\textwidth}
    \centering
    \includegraphics[width=0.48\textwidth]{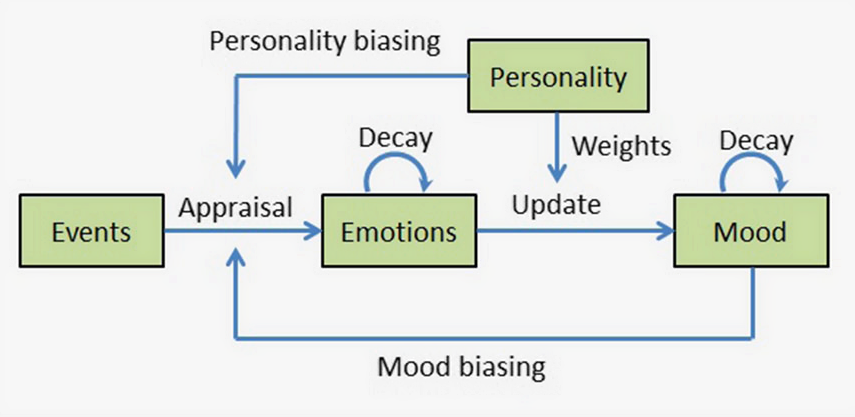}
    \caption{The dynamics between emotions, mood and personality}
    \label{fig:affectdynamics}
\end{wrapfigure}
The last decades saw the development of several systems that attempt to model the dynamics between different affective concepts and particularly between emotions, moods, and personality traits. One of the most popular approaches that was first introduced by Gebhard \cite{Gebhard2005} with his ALMA model and later improved and extended by many other researchers represents affective concepts in a dimensional space of Pleasure-Arousal-Dominance (PAD) and exploits this 3D space to compute and simulate the accumulative and subtractive behaviour that can occur between different affective phenomena. Nadine's affective system follows the same approach and implements sophisticated equations to calculate the dynamics between emotions, mood, and personality more accurately \cite{Zhang2016,Zhang2015a}. Figure \ref{fig:affectdynamics} illustrates the interaction between the different affective concepts. Emotional appraisal in our implementation is determined by the LLM as described in Section \ref{sec:3.3.1}. Both the personality traits that are represented as the Big-Five’s openness, conscientiousness, extraversion, agreeableness, and neuroticism dimensions \cite{McCrae1992}, and emotions that are extracted through the LLM are converted to Mehrabian’s Pleasure-Arousal-Dominance (PAD) dimensional space \cite{Mehrabian1996,Mehrabian1996a,Mehrabian1995}. This 3D space is then used to represent and simulate the interaction between personality, emotions, and moods.

The emotion category and intensity, as well as the robot's personality, are used to update the mood values. The updated mood and the personality of the robot are then used to calculate a more balanced emotional intensity. The idea behind this approach is to generate more natural emotional behaviour, as a person in a certain mood is less likely to have stronger emotions that do not correspond to their mood and vice-versa. A person in an angry mood is more likely to experience intense angry emotions in the case of negative emotions than they would experience intense happy emotions in the case of positive events. The same theory also applies to the personality as a stable construct that influences both the immediate emotions and the longer-lasting moods.

\subsubsection{LLM-RAG based memory and knowledge system}
\label{sec:3.3.3}

\begin{figure*}[h]
 \centering
 \includegraphics[width=0.85\textwidth]{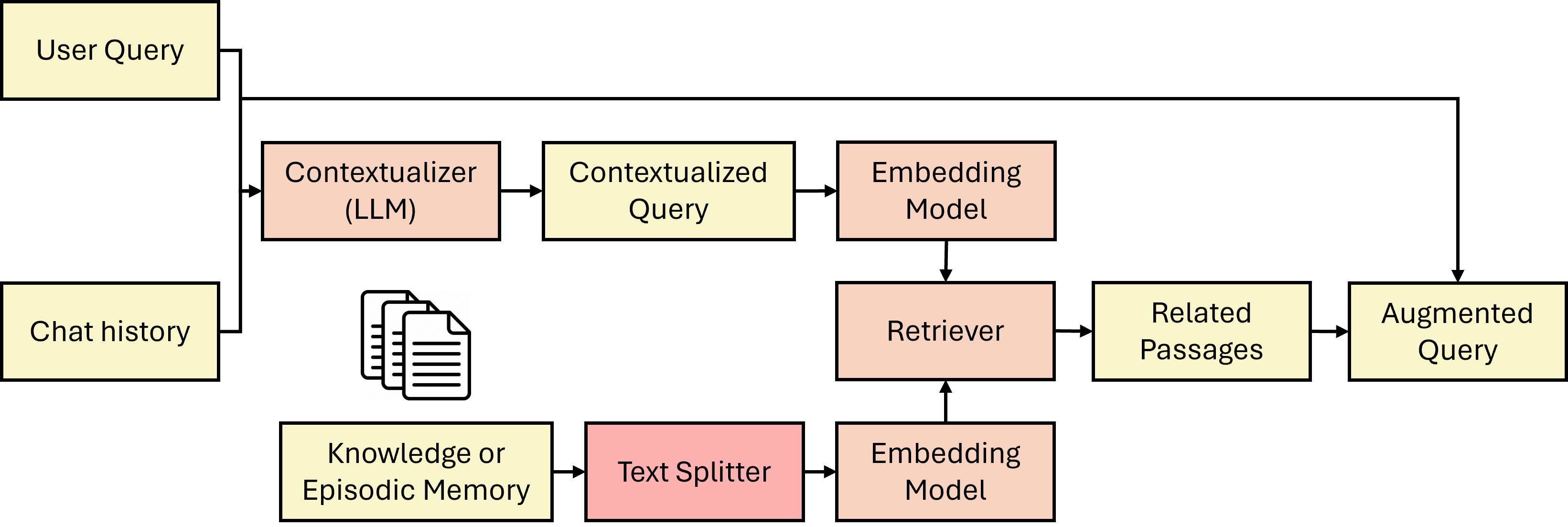}
 \caption{
Our retrieval augmented generation-based memory system workflow
}
\label{fig:rag_workflow}
\end{figure*}

In this section, we demonstrate the RAG system adopted in our interaction module. The RAG framework helps the LLMs generate better responses by referring to external data that is not used during the LLM parameter training process. The RAG approach is especially efficient when LLMs need to be equipped with user-specific information or domain-specific knowledge. Our interaction system employs the power of RAG to increase the quality of the robot’s responses by retrieving the related information from the user's specific long-term episodic memory and specific domain knowledge. The overall workflow of our RAG system is depicted in Figure \ref{fig:rag_workflow}.

Prior to retrieving the passages related to the user query, we store the vector embeddings of knowledge or user-specific episodic memory using an embedding model and a vector database. Through our experiments, we discovered that ‘text-embedding-3-large’ from OpenAI and Chroma DB, an open-source vector database \footnote[3]{\url{https://www.trychroma.com/}}, are effective for the embedding model and the vector database respectively. 

Multiple documents containing the target knowledge are loaded from various sources, and then extensive text in each document is segmented into smaller chunks. This process is crucial for effective retrieval since the embedding model is applied to each chunk. If the chunk size is too small, the embedding may not contain the valuable semantic information of the chunk. On the other hand, if the chunk size is too large, the embedding may also fail to attain important context hidden by high understanding context, which leads to adversely affecting the quality of the retriever results. Our RAG system contains two kinds of memories having different data attributes. The knowledge data is static and the user-specific long-term episodic memory is dynamic. For this reason, we employ different strategies for the text splitter used for knowledge retrieval and episodic memory retrieval as follows:

\noindent \textbf{Knowledge (Static): } the text splitter segments each knowledge document with a fixed chunk size. In our splitter, the chunk size is set to 1,000 with overlapping 200.

\noindent \textbf{Long-term Episodic Memory (Dynamic): } Different to the knowledge data, episodic memory is dynamic and important semantics may be contained in different sizes of text depending on the size of the dialogue. These attributes of episodic memory hinder segmenting the data into a fixed chunk size. We solved this challenge by adopting a sliding window splitter having adaptively varying chunk sizes. The system stores conversations every five back-and-forth interactions (ten messages) as a segment in the user-specific vector space, which serves as a unit in the embedding process. In this process, we store chunks overlapping one interaction so as not to miss contextual or relevant information in the successive dialogue. Through this process, we successfully retrieve the relevant episodic memories for the user.

Once the segmented text is embedded into knowledge vector space or user-specific vector spaces and stored in the vector database, the preparation for the retrieval is complete. Each time the user asks a question, knowledge and episodic memories relevant to this question are retrieved from the RAG system. Firstly, the user query is reformulated by referring to the current chat history (short-term memory) by an LLM. This process helps the RAG system refer to existing context from past messages. For example, if the user asks a follow-up question like ‘Can you elaborate on the second point?’, the system cannot understand the context with only this question, but with the context of the previous interactions it will reformulate the question to convey all relevant information. We employ the contextualiser system template from Langchain \footnote[4]{\url{https://python.langchain.com/docs}}. The contexualised query is transformed into vector embeddings by the embedding model, and the embeddings are utilised to retrieve the relevant passages by comparing the similarities between the embedding vectors of knowledge or episodic memories. Through this process, we retrieve the top-five related passages and augment the user query with the related passages along with the chat history in the SoR-ReAct prompt as described in section \ref{sec:3.3.1}. The augmented query is passed to the dialogue management module and processed to generate the robot’s response.

\subsubsection{LLM Tools}
\label{sec:3.3.4}

In this section, we introduce the tools used in our SoR-ReAct agent and the reasons for the selection of each tool. LLM’s tool-use capabilities enhance the quality of response of LLM agents by enabling the agents to interact with external tools (APIs). We leverage four tools as depicted in Figure \ref{fig:system_architecture}.

\noindent \textbf{Internet Search: } The tool is utilised when the LLM agent encounters queries requiring information from online sources or up-to-date data. The internet search tool enhances the agent’s ability to adapt to novel or evolving topics.

\noindent \textbf{News Search: } The tool is activated to fetch the latest or current events from various news platforms. This feature enables the agent to keep up with real-time news updates, thereby enhancing the timeliness and informed interaction experience. The tool returns summarised news from the retrieved news articles.

\noindent \textbf{Weather Search: } When users inquire about weather conditions or forecasts, the weather search tool is invoked to retrieve meteorological data. By incorporating real-time weather information into its responses, the agent can offer users accurate and location-specific weather forecasts, enhancing user satisfaction. The tool returns summarised weather data, not full weather forecast data.

\noindent \textbf{Wikipedia: } The Wikipedia tool provides the LLM agent with access to a vast repository of encyclopedic knowledge on a wide array of topics. When confronted with inquiries requiring in-depth explanations or background information, the agent leverages this tool to retrieve authoritative and comprehensive content from Wikipedia articles. By harnessing the wealth of information available on Wikipedia, the LLM agent can enhance the depth and quality of its interactions with users.

\subsection{Robot Control Module}
\label{sec:3.4}
The robot control module is responsible for generating the verbal and non-verbal behaviour of the social robot based on the input received from the interaction module. 
It takes information on the current robot's emotions, gestures, pose, gaze, and dialogue utterances and generates actuator-driven robot animations and synthesised speech.

The control model consists of the following components:
\begin{itemize}
    \item The \textit{\textbf{speech synthesis}} component takes dialogue utterances and generates the robot speech and the corresponding visemes for lip synchronisation.
    \item The \textbf{\textit{lip synchronisation}} takes the visemes from the speech synthesis component and generates high-level lip animations for the animation engine.
    \item The \textbf{\textit{facial expression}} component takes the emotion value from the interaction module and generates high-level face animations for the animation engine.
    \item The \textbf{\textit{gesture and pose}} component generates gesture and pose animations based on input from the interaction module, as well as automatic animations for realistic idle behaviour. The generated high-level animations are subsequently sent to the animation engine.
    \item The \textbf{\textit{gaze}} component takes the user locations perceived by the perception module and generates realistic high-level look-at and look-away gaze animations for the animation engine.
    \item The \textbf{\textit{animation engine}} takes the high-level animations generated by the components mentioned above and generates an animation timeline where it intelligently blends animations together and creates transitions where needed. Subsequently, the animation engine iterates the animation timeline each time interval and sends the corresponding frame to the actuator control component.
    \item The \textbf{\textit{actuator control}} component receives animation frames from the animation engine at each interface and calculates the corresponding actuator values for each facial and bodily joint of the robot. These actuator values are subsequently sent to the robot via serial communication.
\end{itemize}

\section{Results}
\label{sec:4}

Ablation studies were conducted to assess the efficacy of each component within SoR-ReAct. The utility of each component is elucidated through qualitative examples in the subsequent sections.

\subsection{Tool use}
\label{sec:4.1}

\begin{figure}[h]
    \centering
    \begin{subfigure}[b]{0.4\textwidth}
        \centering
        \includegraphics[width=\textwidth]{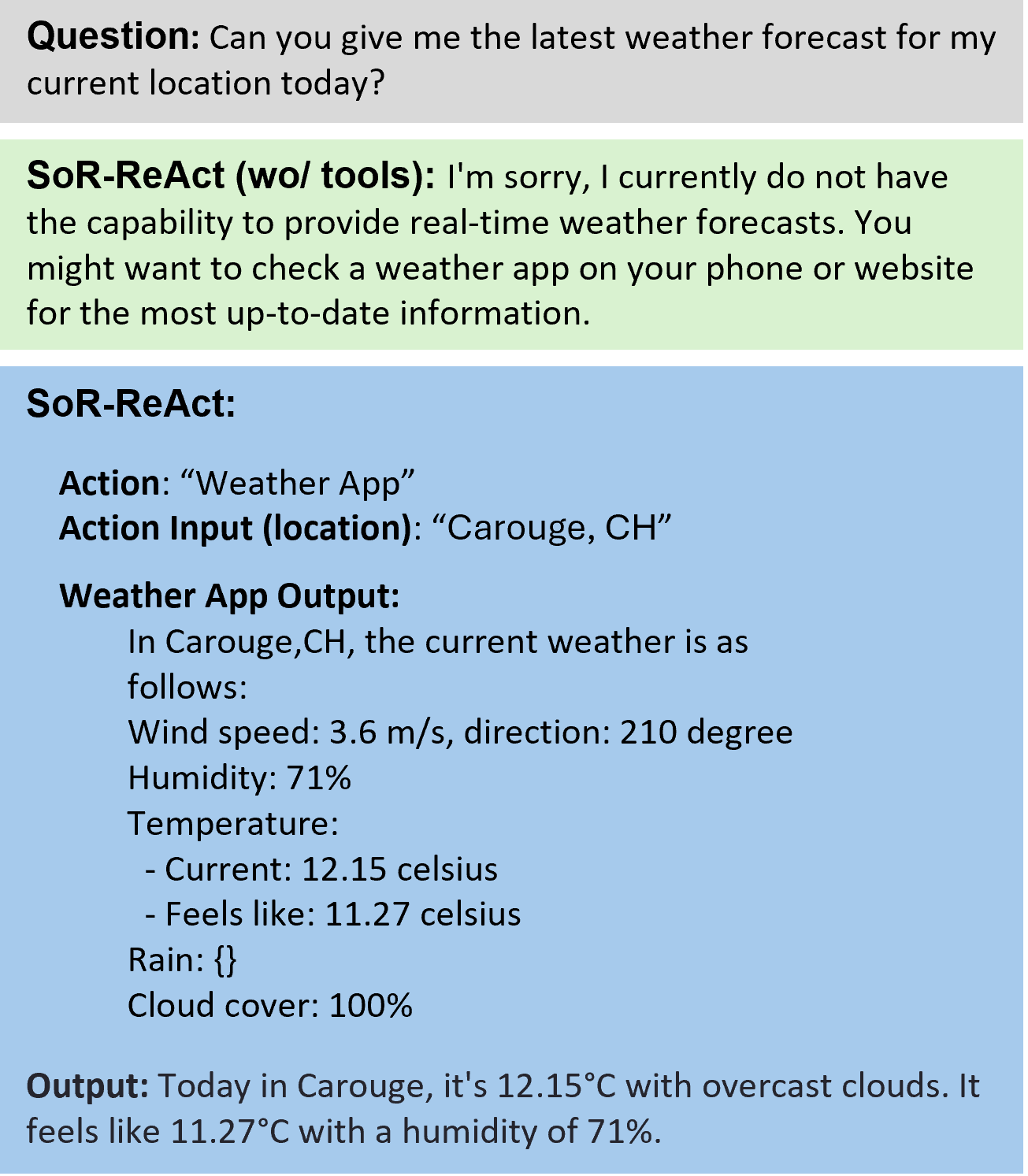}
        \caption{ }
        \label{fig:exp_tool_weather}
    \end{subfigure}
    \begin{subfigure}[b]{0.4\textwidth}
        \centering
        \includegraphics[width=\textwidth]{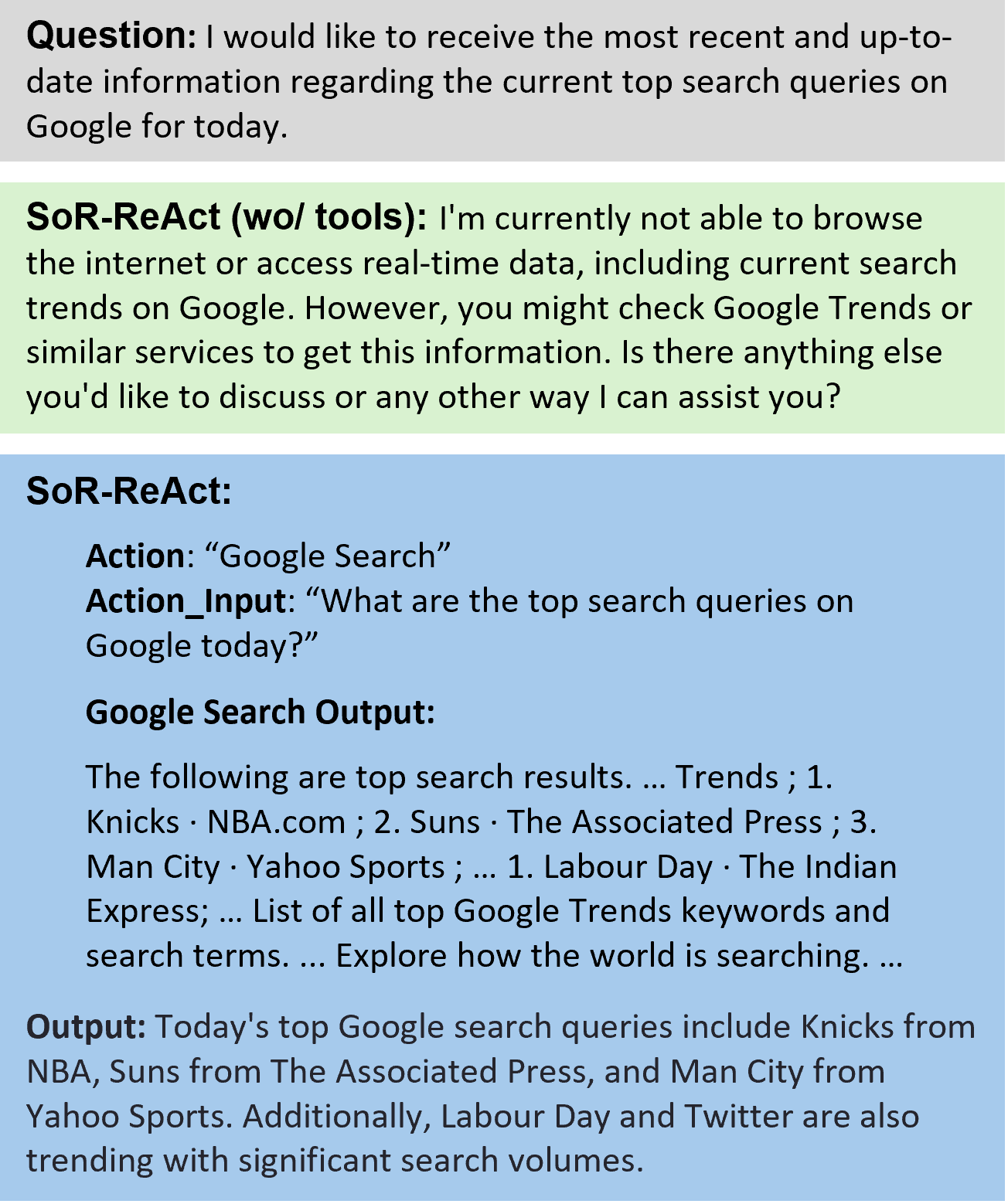}
        \caption{ }
        \label{fig:exp_tool_search}
    \end{subfigure}
    \caption{Ablation of the tool use capability of SoR-ReAct (a): Weather app, (b): Google search}
\end{figure}

As delineated in section \ref{sec:3.3.4}, we have integrated various tools into our agent to facilitate interaction with external resources, thereby augmenting response quality. Evaluation of tool use capability entailed a comparative analysis between two settings: one with access to tools and the other without. Testing encompasses the utilisation of weather search and internet search tools. The results, illustrated in figure \ref{fig:exp_tool_weather} and \ref{fig:exp_tool_search}, demonstrate that the SoR-ReAct agent endowed with tool utilisation capability can provide users with real-time data such as weather forecasts and current search trends. This capability not only enriches the user experience but also underscores the versatility and adaptability of the SoR system in engaging with dynamic information environments.

\subsection{Affective system}
\label{sec:4.2}

\begin{wrapfigure}{r}{0.5\textwidth}
    \centering
    \includegraphics[width=0.38\textwidth]{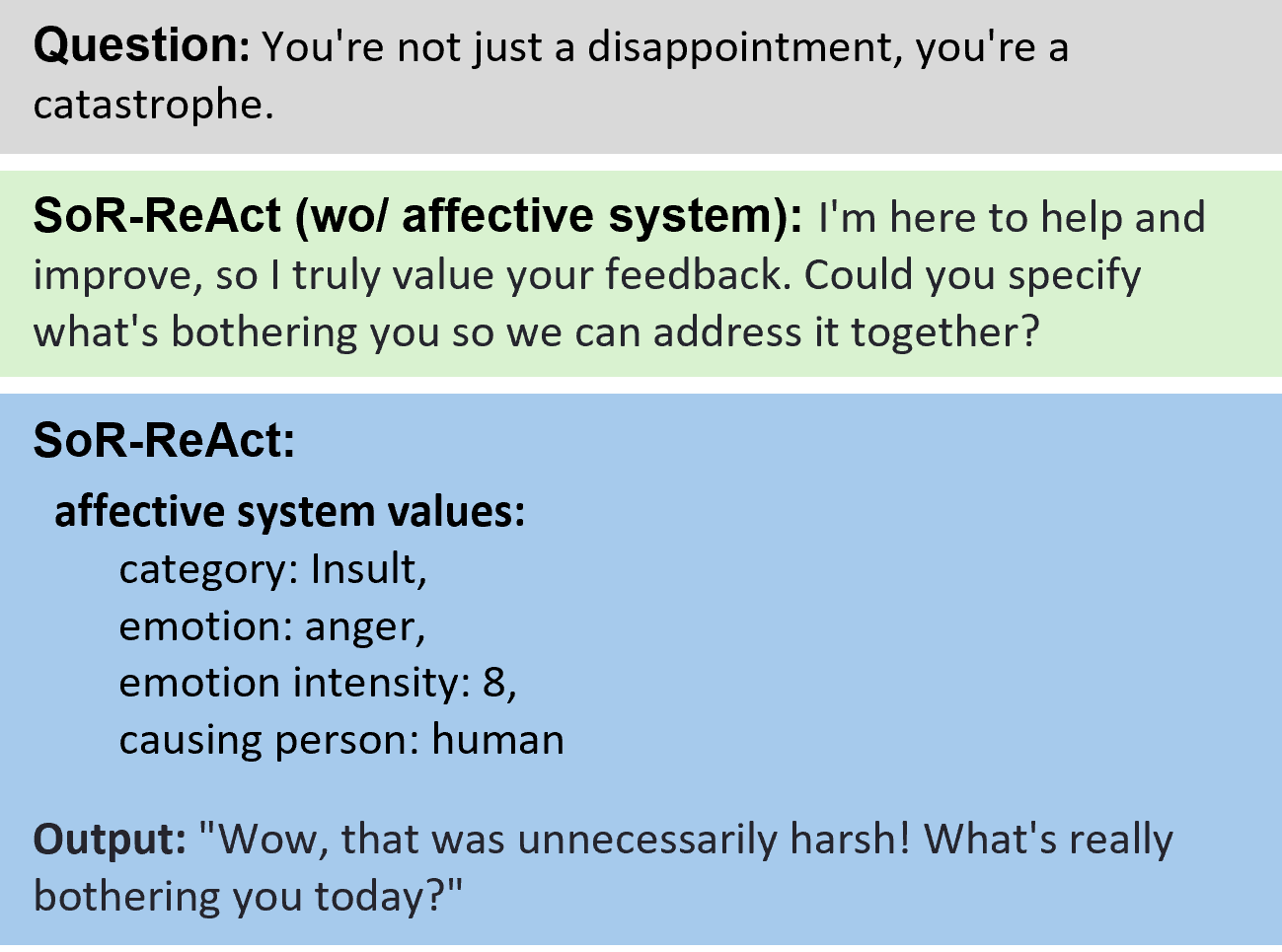}
    \caption{Ablation of the affective system of SoR-ReAct.}
    \label{fig:exp_affective}
\end{wrapfigure}

Our SoR-ReAct agent incorporates an affective system designed to simulate internal emotional states, as explained in section \ref{sec:3.3.2}. These emotional states influence the robot's behaviour, including its responses and gestures, thus enhancing its capacity for empathetic interactions with users. In this section, we evaluate the efficacy of the affective system through a comparison of responses generated by agents with and without the aforementioned system. As depicted in figure \ref{fig:exp_affective}, the agent equipped with the affective system produces responses according to human-like emotional nuances, with the aim of fostering deeper rapport and understanding in human-robot interactions. Expected outcomes include smoother communication facilitated by nuanced emotional expression and fostering trust and satisfaction, which are crucial elements in the development of social robots.


\subsection{Long-term memory}
\label{sec:4.3}

\begin{figure*}[h]
 \includegraphics[width=\textwidth]{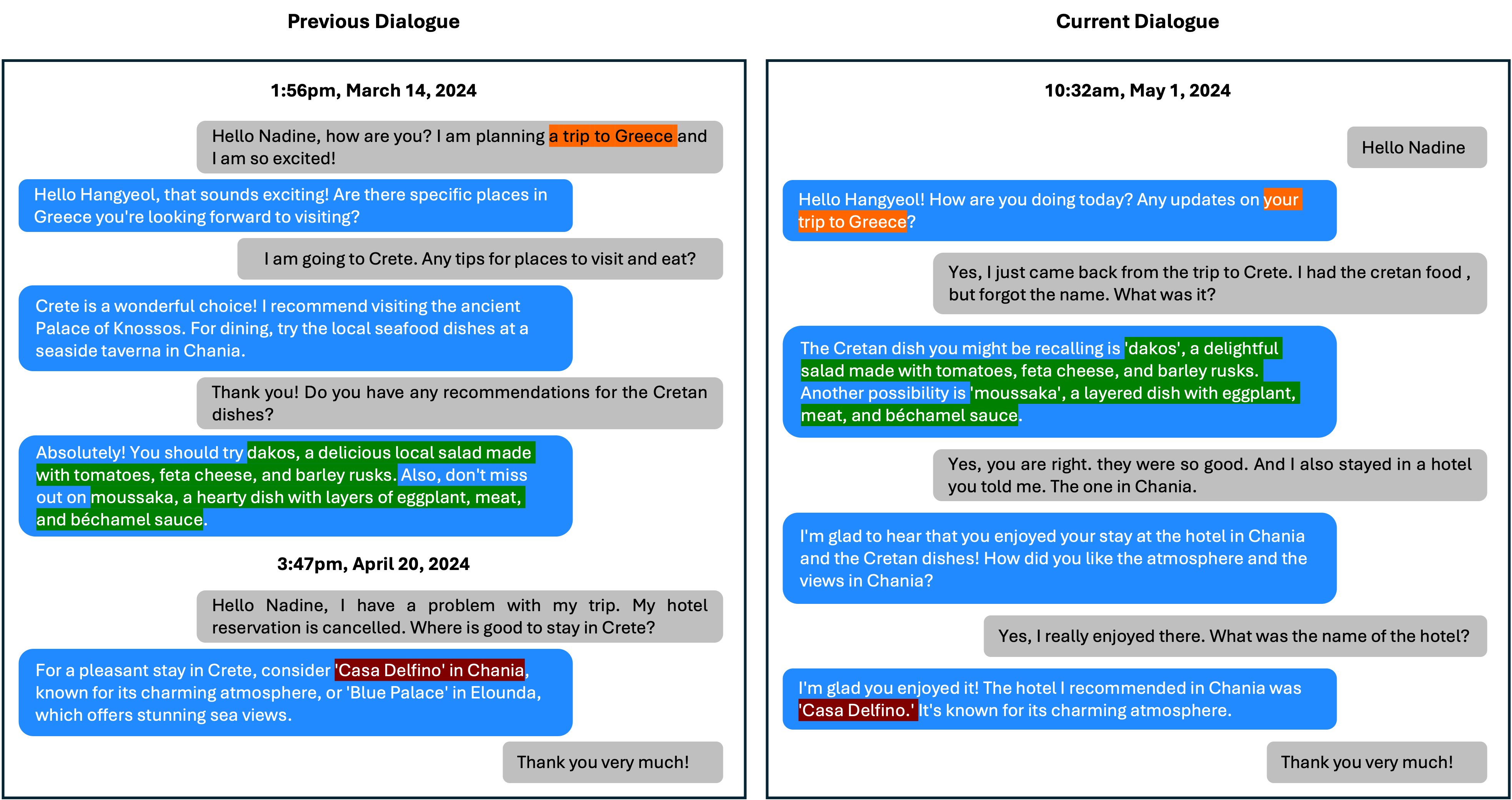}
 \caption{A long-term memory capability example.}
\label{fig:exp_memory}
\end{figure*}

Another pivotal aspect of social robot agents is long-term memory capability. Herein, we illustrate the long-term memory proficiency of our agent, as detailed in section \ref{sec:3.3.3}. Figure \ref{fig:exp_memory} showcases the agent's capacity to retrieve pertinent information from prior dialogues, enabling it to generate contextually appropriate responses based on stored memory. This functionality empowers the robot to remember users across sessions, fostering deeper rapport and continuity in interactions. By seamlessly integrating past interactions into present dialogue, the SoR system demonstrates an advanced level of cognitive adaptability.

\section{Conclusion}
\label{sec:5}

In this study, we presented an LLM-driven robotic system tailored for social robots, comprising three pivotal modules: the perception module, the interaction module, and the robot control module. Our research centers on enhancing the interaction module to enable the robot to mimic more human-like behaviours, achieved by attaining versatile capabilities such as external knowledge access, long-term memory for recognised users, and sophisticated emotional appraisal. The seamless integration of the aforementioned features is facilitated by our SoR-ReAct agent framework. We have demonstrated the efficacy of our robotic system through ablation studies of each component in the SoR-ReAct agent. It is worth noting that our system is designed for a single-party interaction where only one user interacts with the robot at a time. Future work includes extending its capabilities to detect active speakers, handle multi-party interactions with more users interacting with the robot simultaneously, and elucidate the dynamics of user interactions within a group setting.

\section{Acknowledgements}
Co-funded by the European Union. Views and opinions expressed are however those of the author(s) only and do not necessarily reflect those of the European Union. Neither the European Union nor the granting authority can be held responsible for them. 

This work has received funding from the Swiss State Secretariat for Education, Research and Innovation (SERI). This work was performed within the frame of Horizon Europe project INDUX-R (101135556) and IDS (100.133 IP-ICT).

\bibliographystyle{unsrt}  
\bibliography{NadinePlatform}

\end{document}